\newcommand{\CLASSINPUTinnersidemargin}{18mm}
\newcommand{\CLASSINPUToutersidemargin}{12mm}
\newcommand{\CLASSINPUTtoptextmargin}{20mm}
\newcommand{\CLASSINPUTbottomtextmargin}{25mm}
\documentclass[10pt,conference,a4paper]{IEEEtran}

\usepackage[utf8]{inputenc}
\usepackage{dblfloatfix}
\usepackage{float}
\usepackage{amsmath}

\usepackage{times}
\usepackage{comment}

\usepackage{graphicx}
\usepackage{subfigure}
\DeclareGraphicsExtensions{.png,.eps,.ps,.pdf}

\usepackage{url}

\usepackage[english]{babel}

\begin{document}

\title{Classification of Industrial Control Systems screenshots using Transfer Learning}


\makeatletter
    \newcommand{\linebreakand}{%
      \end{@IEEEauthorhalign}
      \hfill\mbox{}\par
      \mbox{}\hfill\begin{@IEEEauthorhalign}
    }
\makeatother

\author{\IEEEauthorblockN{\small Pablo Blanco-Medina}
\IEEEauthorblockA{\small
Dept. IESA.\\
Universidad de León\\
Researcher at INCIBE\\
pblanm@unileon.es}
\and
\IEEEauthorblockN{\small Eduardo Fidalgo}
\IEEEauthorblockA{\small
Dept. IESA.\\
Universidad de León\\
Researcher at INCIBE\\
eduardo.fidalgo@unileon.es}
\and
\IEEEauthorblockN{\small Enrique Alegre}
\IEEEauthorblockA{\small
Dept. IESA.\\
Universidad de León\\
Researcher at INCIBE\\
enrique.alegre@unileon.es}
\and
\IEEEauthorblockN{\small Francisco Jáñez-Martino}
\IEEEauthorblockA{\small
Dept. IESA.\\
Universidad de León\\
Researcher at INCIBE\\
fjanm@unileon.es}
\linebreakand
\IEEEauthorblockN{\small Roberto A. Vasco-Carofilis}
\IEEEauthorblockA{\small
Dept. IESA.\\
Universidad de León\\
Researcher at INCIBE\\
rvasc@unileon.es}
\and
\IEEEauthorblockN{\small Víctor Fidalgo Villar}
\IEEEauthorblockA{\small 
\small Spanish National \\ Cybersecurity \\ Institute (INCIBE)\\
victor.fidalgo@incibe.es}}

\maketitle

\begin{abstract}
Industrial Control Systems depend heavily on security and monitoring protocols. Several tools are available for this purpose, which scout vulnerabilities and take screenshots from various control panels for later analysis. However, they do not adequately classify images into specific control groups, which can difficult operations performed by manual operators.
In order to solve this problem, we use transfer learning with five CNN architectures, pre-trained on Imagenet, to determine which one best classifies screenshots obtained from Industrial Controls Systems. Using $337$ manually labeled images, we train these architectures and study their performance both in accuracy and CPU and GPU time. We find out that MobilenetV1 is the best architecture based on its $97,95$\% of F1-Score, and its speed on CPU with $0.47$ seconds per image. In systems where time is critical and GPU is available, VGG16 is preferable because it takes $0.04$ seconds to process images, but dropping performance to $87,67$\%.

\end{abstract}


{\bf Type of contribution:}  {\it Original research}

\section{Introduction}

Interconnection between electronic devices that are connected to the Internet has become a necessity, ensuring the control, communication and monitoring of multiple systems. Those systems that are exposed online should be deployed under various security measures to avoid potential attacks \cite{wolf2017safety}.

In critical infrastructures, such as healthcare, transportation or manufacturing, a system shutdown or restart would lead to severe economic and social consequences, as well as significant time costs. For this reason, these systems must rely on constant monitoring \cite{wolf2017safety}.
Additionally, the threat of a potential security breach can rank from information leak to system overtake, which entails high risks in environments such as Industrial Control Systems (ICS) \cite{wolf2017safety}.

Supervisory Control And Data Acquisition (SCADA) systems, used to control both physical equipment and ICS infrastructure, are commonly referred to as Operational Technology (OT) systems, which directly control and monitor specific devices. Other industrial systems used to control software, including management, storage and delivery of data, are known as Information Technology (IT) systems. \cite{ITVSOT-conklin2016vs}. Fig. \ref{fig:figITOT} presents a screenshot of an IT and an OT system, with the purpose of appreciating their differences.

\begin{figure}[t]
\centerline{
\includegraphics[width=\linewidth, scale=0.5]{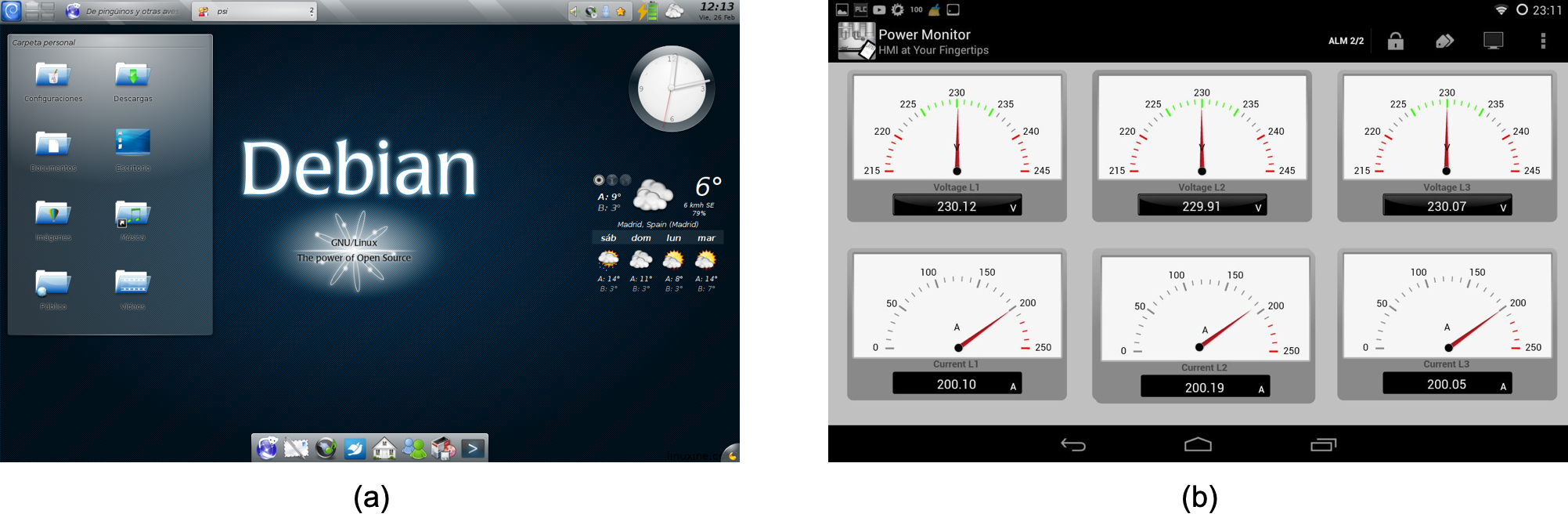}
}
\caption{IT (a) and OT (b) sample images. IT focuses on software management and data control, while OT systems help directly monitor device values.}
\label{fig:figITOT}
\end{figure}

To monitor these exposed assets, Law Enforcement Agencies (LEAs) use Open Source INTelligence (OSINT) tools \cite{OSINT-2016}. 
In particular, specialized tools , such as Shodan \cite{shodan-genge2016shovat}, known as metasearchers, monitor the open ports of a network, as well as the services of the devices that are exposed to Internet. For those services that include a Graphical User Interface (GUI), specific metasearchers usually take screenshots to log relevant information graphically. The classification of these assets is very useful to determine the type of compromised devices and, therefore, the taken screenshots help to discover vulnerabilities, to classify the devices based on the images taken and to analyse the obtained information posteriorly. 

However, these metasearchers may not correctly classify images as belonging to SCADA or ICS systems what frequently makes necessary a manual classification. Due to the large number of devices connected to the Internet and the multiple monitoring options existing in the metasearchers, this manual process can be an arduous task for a human operator. Moreover, the changing environment and continual updates of these systems increase the difficulty of classifying these images using traditional methods such as hand-crafted features.

In order to solve this problem, we propose the use of Deep Learning to automatically classify the screenshots taken during the monitoring of open ports and devices exposed to the Internet. We used pre-trained Convolutional Neural Networks (CNNs) alongside transfer learning to build an image classifier that labels ICS images in three categories: \textit{IT}, \textit{OT} and \textit{Others}.

Our proposal helps enhance the analysis and classification of graphical interfaces linked to devices, giving a greater context to the security analyst who monitors them and allowing quicker action in attack cases such as ransomware or value alteration.

The rest of the paper is organized as follows. Section~\ref{sec:sota} presents a summary of state-of-the-art in image classification approaches and architectures. In Section~\ref{sec:methodology}, we introduce the methodology followed. Section~\ref{sec:results} discusses our experimental settings and the obtained results. Lastly, in Section~\ref{sec:conclutions}, we present our conclusions and future lines of work.

\section{State of the Art} \label{sec:sota}

Image classification can be defined as the task of assigning a label to an image, which indicates that the image belongs to the category represented by that label \cite{rawat2017deep}. Traditionally, to carry out this task, hand-crafted features \cite{fidalgo2019fusion} were extracted from the images and then used to train the classifiers.

Since their breakthrough by achieving the best result on the ILSVRC (ImageNet Large Scale Visual Recognition Challenge) \cite{russakovsky2015imagenet}, CNNs have established themselves amongst the best image-based learning algorithms \cite{rawat2017deep}. Despite this, there are cases such as lack of image dataset or challenging classification tasks in which manually-crafted feature extraction can outperform the results obtained by CNNs \cite{fidalgo2018boosting,fidalgo2019classifying}.




CNNs have improved their performance over the years by optimizing their parameters and changing their structures to fit different problems \cite{rawat2017deep}. Their progress has also been possible due to technological advances, such as the use of Graphics Processing Units (GPUs).

As seen in \cite{cnn-khan2019survey}, CNNs can be divided into $7$ different categories according to their architecture; spatial exploitation, depth, multi-path, width, feature-map exploitation, channel boosting and attention. Multiple networks can appear in various categories.






In order to train these networks, a large amount of data is needed. Data-gathering and annotation can be a complex and time consuming process. Furthermore, these datasets may soon become outdated, needing the addition of new data \cite{transfer-survey}.

Transfer learning is a technique that takes a model trained for a specific problem and applies it to a similarly related task \cite{hussain2018study}. This approach is used to retain the features obtained from bigger datasets and use them to train a new model on a smaller, similar dataset \cite{hussain2018study}.

Several works have studied the use of transfer learning applied to CNNs for the task of image classification. Hussain et al. (2018) \cite{hussain2018study} studied the application of transfer learning on the InceptionV3 \cite{inceptionv3} architecture pre-trained on the ImageNet dataset \cite{imagenet} and re-trains it on the CIFAR-10 \cite{cifar-10} dataset, obtaining $70.1$\% accuracy and surpassing CNNs trained from scratch on this dataset.

Sharma et al. (2019) \cite{sharma2018analysis} apply transfer learning on AlexNet \cite{alexnet}, GoogLeNet \cite{googlenet} and ResNet50 \cite{ResNet50}. For each network, they replace the last three layers with a fully connected layer, a softmax layer and a classification output layer. Afterwards, they train the networks on the CIFAR-10 dataset, obtaining classification accuracy per image category. The resulting average performance for each network was $36.12$\% for AlexNet, $71.67$\% for GoogLeNet and $78.10$\% for ResNet50.

Extensive architecture research and engineering are required to improve neural network classification. To fix this problem, Bello et al. (2017) \cite{bello2017neural} proposed an approach called Neural Architecture Search that helps optimize architecture configurations, improving classification performance and training time on the CIFAR-10 dataset.

However, this technique has a high computational cost when training the architectures on large datasets, such as Imagenet. Therefore, \cite{zoph2018learning} proposed the use of a smaller dataset,  CIFAR-10, as a proxy and then transfer the learned architecture to the Imagenet dataset.

The resulting architecture, called NASNet, is compared to multiple CNNs such as MobileNet-224 \cite{mobilenetv1}, Inception-ResnetV2 \cite{inception-resnet-szegedy2017inception} and Xception \cite{xception} on the task of image classification on the imagenet dataset, comparing both the number of parameters and the accuracy. The proposed solution achieved state-of-the-art results on the Imagenet dataset.

\section{Methodology} \label{sec:methodology}

For our proposal, we used a dataset of ICS images provided by INCIBE. It contains a total of $337$ manually labeled images of varying sizes, which were retrieved using multiple metasearchers. This dataset is split into $74$ IT images and $263$ OT images, all of which are used for training and testing our system.

We decided to implement transfer learning instead of re-training the architectures from scratch, due to the limited amount of images available. 
For the experiments, we selected five architectures to apply to our image classification problem; InceptionV3 \cite{inceptionv3}, MobilenetV1 \cite{mobilenetv1}, ResNet50 \cite{ResNet50}, VGG16 \cite{vgg16} and Xception \cite{xception}.

We chose these architectures because they are commonly used in similar problems that imply transfer learning  \cite{hussain2018study,sharma2018analysis,zoph2018learning}. Additionally, MobilenetV1 was chosen due to the real-time based nature of the given task, with its focus on mobile, lightweight deployment.

\begin{figure*}[t]
  \includegraphics[width=\textwidth]{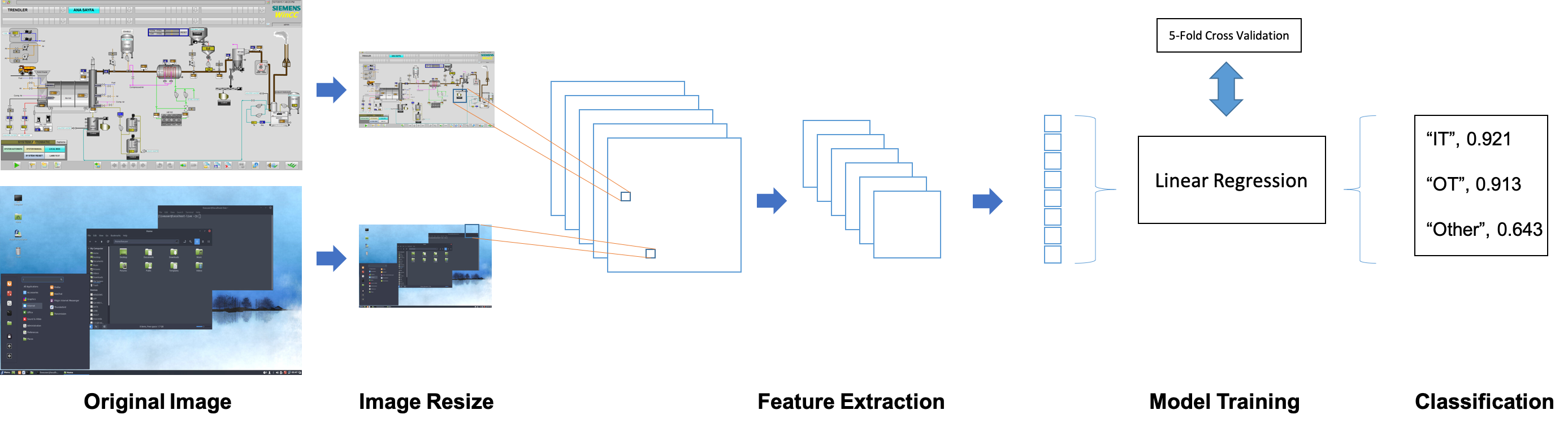}
\caption{Methodology pipeline. First, the input image is resized to fit each architecture required dimensions. After loading the CNN, the final layer is frozen for feature extraction. These features are used to train a linear regression classifier on GPU, using 5-Fold Cross Validation. After training, the classifier outputs a label (IT, OT or Others) and the highest confidence score.}
\label{fig:figTRANSFER}
\end{figure*}

Fig. \ref{fig:figTRANSFER} presents an overview of the proposed system for classifying screenshots. 
For each network, we freeze the final layer using the ImageNet pre-trained weights, and obtain the features before classification. Then, we use those features to train a Logistic Regression model.

Using the trained classifiers, we label images in  three categories: \textit{IT}, \textit{OT} and \textit{Others}, according to the classifier's confidence score. If it is below a certain threshold, images are classified as \textit{Others}. This third category helps identify additional clusters retreived from the metasearchers, such as IoT images, that could be useful for LEAs to add as future labels.

\section{Experimental Results and Discussion} \label{sec:results}

\subsection{Experimental Settings}

We evaluated our proposal on an Intel Xeon E5 v3 computer with $128$GB of RAM using an NVIDIA Titan Xp GPU for both training and testing.

All of the five CNN architectures are implemented using Python3 under the Keras library \cite{chollet2015kera} with Theano as the backend. The logistic regression classifier is implemented using the Scikit-learn Python library \cite{scikit-learn}.

To fit the architecture's input size, each image is resized to the required value. For VGG16 and ResNet50, images are fixed to a size of 224x224, while for MobilenetV1, Xception and InceptionV3 they are scaled to 299x299. We feed the resized images to these pre-trained networks, extract the features and train our classifier using them.

Due to the difference between the available images per category, we implement 5-Fold Cross Validation, generating five models per architecture. This technique helps reduce model bias when compared to other approaches such as the train-test split. The images are split into five folds, four of which are used to fit the model, while the last one is used for validation. This process is repeated until every fold has been used to test the proposed model.

To measure the performance of these models, we use the F1 score, as can be seen in Eq. \ref{eq:eqF}. We chose this metric to better represent the robustness of our classifier, as it is the harmonic mean of the Precision and Recall measures, detailed in Eq. \ref{eq:eqP} and Eq. \ref{eq:eqR} respectively. In these equations, TP and FP represent True and False Positives, respectively, while FN indicates False Negatives.

\begin{equation}
  \begin{aligned}
    P = \frac{TP}{TP+FP}.
  \end{aligned}
\label{eq:eqP}
\end{equation}

\begin{equation}
  \begin{aligned}
    R = \frac{TP}{TP+FN}.
  \end{aligned}
\label{eq:eqR}
\end{equation}

\begin{equation}
  \begin{aligned}
    F1 = 2 \cdot \frac{P \cdot R}{P + R}.
  \end{aligned}
\label{eq:eqF}
\end{equation}

In the case of image classification, a true positive is considered when an image is assigned its correct label. For a particular class, a false positive is considered for each image that has been labeled as belonging to said class, but belongs to a different one. At the same time, false negatives are considered as the images from the class that have been assigned other labels.

After finishing the training, we select one of the five generated models per architecture at random to obtain the F1 score result. We also measure the time needed for feature extraction and classification in both CPU and GPU. We retrieve both the mean and standard deviation in each task.

\subsection{Results Discussion}

The results of our experiments can be seen in Table \ref{tab:tab_transfer}. The image classifier built using MobilenetV1 scored the highest on the given images, with an F1-Score of $97.95$\% and a variance of $1.08$\%. InceptionV3 scored the second-best performance with $96.66$\%, but at a much higher variance of $2.68$\%. 

Both approaches scored more than $10\%$ over the rest of the methods, with the lowest results obtained by ResNet50 and VGG16. Although these architectures obtained the less F1-Score variance at $0.35$\% using 5-fold cross-validation, which verifies their stability, the score difference is high enough to dismiss when choosing amongst the architectures for real-time applications.

Regarding real-time performance, except in the particular case of ResNet50, all of the architectures show significant improvement in image processing time when comparing GPU to CPU, being over $3$ times faster on average.

In CPU, MobilenetV1 obtained the best result with $0.47$ seconds per image, surpassing ResNet50's $0.77$ seconds. Although slower on average, VGG16 obtained the least deviation on the test images, with $0.16$ seconds over MobilenetV1's $0.59$ second variance.

Finally, in GPU, 
VGG16 obtained the fastest time per image in GPU with $0.04$ seconds per image, three times faster than MobilenetV1 and with the least time variance ($0.16$ seconds) across all architectures. 

Given the real-time processing issue of monitoring ICS systems, as well as its particular focus on light deployment, MobilenetV1 should be considered as the first solution due to having the highest average F1 score as well as much faster CPU and GPU average times than the rest of the architectures. InceptionV3 scores close to MobilenetV1, but is twice as slow in CPU and has a higher time deviation.

\begin{table}[hbt!]
\centering
\caption{Transfer Learning Results on our custom dataset.}
\label{tab:tab_transfer}
\resizebox{\linewidth}{!}{
\begin{tabular}{c c c c}
\hline
Architecture & F1-Score & CPU (s) & GPU (s)\\\hline 
InceptionV3 & $96.66$\% (+/- $2.68$\%) 
& $1.31$s (+/- $1.54$s) 
& $0.41$s (+/- $0.41$s)\\
MobilenetV1 & $\mathbf{97.95}$\textbf{\%} (+/- $1.08$\%)
& $\mathbf{0.47}$\textbf{s} (+/- $0.59$s)
& $0.13$s (+/- $0.91$s) \\
ResNet50 & $87.67$\%
\textbf{(+/- }$\mathbf{0.35}$\textbf{\%)}
& $0.77$s (+/- $1.38$s)
& $1.05$s (+/- $1.57$s)\\
VGG16 & $87.67$\%
\textbf{(+/- }$\mathbf{0.35}$\textbf{\%)}
& $1.07$s \textbf{(+/- }$\mathbf{0.16}$\textbf{s)}
& $\mathbf{0.04}$\textbf{s (+/- }$\mathbf{0.28}$\textbf{s)} \\
Xception & $89.63$\% (+/- $1.51$\%)
& $1.55$s (+/- $0.81$s)
& $0.18$s (+/- $1.42$s) \\\hline
\end{tabular}
}
\end{table}

\section{Conclusions} \label{sec:conclutions}

In this paper, we have presented a transfer learning-based approach for the task of classifying images as belonging to IT or OT systems, to help LEAs with the analysis of ICS images in order to detect and prevent potential security breaches.

We have analyzed five different CNN architectures, using their pre-trained weights on the Imagenet dataset to train a logistic regression image classifier. We validate our approach using F1-Score as well as 5-Fold cross-validation during training.

We trained and tested the architectures on a $337$ image dataset provided by INCIBE, containing $74$ IT images and $263$ OT images, registering the average time in the classification of each image, both in CPU and GPU.

Our results show that the best CNN architectures for this problem are MobilenetV1 and InceptionV3, achieving $97.95$\% and $96.66$\% F1-Score respectively on the given dataset. Furthermore, MobilenetV1 is the best architecture in CPU time, scoring $0.47$ seconds per image. In GPU, VGG-16 obtains a higher speed than MobilenetV1 with a mean time of $0.04$ seconds.

Although faster in GPU, VGG16 scored lower than the rest of the architectures with an F1-Score of $87.67$\%. However, it also obtained the small deviation both in performance ($0.35$\% against MobilenetV1's $1.08$\%) and time ($0.04$ seconds vs $0.13$) needed to process the images.

Our future work will be focused on fine-tuning the proposed solution, adding new layers on top of the given architectures to improve these results. Another possibility is to further extend the current study to include architectures such as VGG19 \cite{vgg16} and InceptionResnetV2 \cite{inception-resnet-szegedy2017inception}.

Additionally, the use of data augmentation techniques to increase the training samples, as well as the inclusion of more detailed classes, are possible future lines of investigation for this image classification task.

Finally, after classification is performed, images can be further analyzed by retrieving information found within them \cite{fidalgo2019fusion}, such as brand or company names using Text Spotting \cite{ICDP2019Blanco,JJAA2019Blanco}. Other approaches can also be applied to help LEAs detect potentially malicious activity \cite{al2017classifying}.

\section*{Acknowledgements}

This work was supported by the framework agreement between the Universidad de Le\'{o}n and INCIBE (Spanish National Cybersecurity Institute) under Addendum 01.
We acknowledge NVIDIA Corporation with the donation of the TITAN Xp and Tesla K40 GPUs used for this research.

\bibliographystyle{IEEEtran}
\bibliography{bibitex}


\end{document}